\documentclass{article}

\usepackage[nonatbib]{nips_2016}

\usepackage{nips_2016}

\usepackage[utf8]{inputenc} % allow utf-8 input
\usepackage[T1]{fontenc}    % use 8-bit T1 fonts
\usepackage{hyperref}       % hyperlinks
\usepackage{url}            % simple URL typesetting
\usepackage{booktabs}       % professional-quality tables
\usepackage{amsfonts}       % blackboard math symbols
\usepackage{nicefrac}       % compact symbols for 1/2, etc.
\usepackage{microtype}      % microtypography
\usepackage{wrapfig}
\usepackage[pdftex]{graphicx}

\title{Learning a CNN-based End-to-End Controller for a Formula Racecar}

\begin{document}

\maketitle

\section{Introduction}
Convolutional neural networks (CNNs) have had much success in the past decade on a many complex vision tasks, successfully passing human benchmarks on object recognition, image segmentation, and video classification \cite{scenelabel, largescale, videoclass}. Much of the success of CNNs can be attributed to their ability to automatically learn feature maps and scale with large datasets  \cite{bigdata}. More recently, deep CNNs have been applied to learn control algorithms associated with steering a regular street-legal sedan \cite{nvidia}. In this report, we aim to adapt the work of Bojarski et al. for the specific purpose of learning a complete drive controller (steering, brake, and throttle) for a Formula SAE racecar navigating a pre-defined cone-delineated track setup.

In addition to optimizing for model accuracy (defined with respect to the actions of model drivers), we aimed to develop CNNs with low pass-through latency and correct behavior in fringe cases (to avoid compounding controller errors). While we follow similar methodologies, note that our problem context differs significantly from the generalized autonomous driving setup in \cite{nvidia}: we are learning controls specific to the MIT Motorsports 2016 racecar on a pre-defined cone setup, targeted for the
2017 Formula Student Germany Driverless (FSGD) competition \cite{fsg, mitmotorsports}. The rest of this paper is organized as follows. In section \ref{setup}, we explain our data collection and software tools. In section \ref{experiments}, we describe our model architectures, loss metrics, experiments, and results for three tasks: discretized steering, real-value steering, and brake/throttle prediction. Finally in section \ref{conclusion}, we discuss our results and present 
directions for future work.

\section{Hardware Setup, Data Collection, and Software Tools}
\subsection{Hardware Setup and Data Collection}
\label{setup}
We collected brake, throttle, steering, and camera footage from a sensor-outfitted Formula SAE racecar. Driving data was collected from runs along a 400-meter orange cone-delineated track provided by FSGD. An outline of the track is viewable at \cite{outline}.

We used a GoPro Hero3 mounted on the car’s top frame to collect 56 minutes of 1920 x 1080 30FPS video. The footage was clipped to exclude frames during which the car was stopped or driving off-track, yielding 28 minutes and 24 seconds of clean driving footage \cite{rawfootage}. Frames were cropped to exclude non-informative peripheral pixels and then resized to 256 x 256 to reduce dimensionality, reduce the parameter count in our network, and circumvent memory limitations during training \cite{cropped}.

\begin{wrapfigure}{l}{0.3\textwidth}
  \centering
  \includegraphics[width=2in]{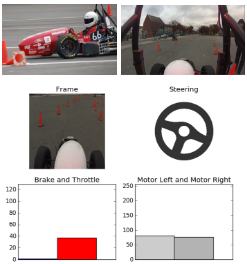}
  \caption{\emph{Top left: the FSAE racecar navigates our cone-delineated path. Top right: sample frame from the GoPro mounted on the car. Bottom: frame from our visual simulator with the cropped video frame and corresponding driving state (section \ref{sim}).}}
  \label{fig:intro}
\end{wrapfigure}
Modifications were made to the car’s central control unit to communicate the driving data every 125 milliseconds over FTDI/UART. We collected steering, brake, and throttle values of three model human drivers, with varying degrees of adherence to the optimal center-line path along the track \cite{logs}. Brake and throttle values ranged from 0 to 256. Collected left and right motor speeds ranged from 0 to 20,000 (in no meaningful units) and were scaled [0, 256]. Steering values were scaled to
[-90, 90] for visualization purposes (interpretable as the degree of the steering wheel).

We batched vehicle measurements into timestamped ‘data frames’:  \texttt{[timestamp, steering, brake, throttle, left\_motor\_speed, right\_motor\_speed]}. Then, using various synchronization signals between the log and the video, we paired each data frame to the video frame closest in time to the data frame timestamp. This resulted in a total of 12097 data/image frame pairs \cite{rawdata}. Using a random 60/20/20 split, we divided the data into 7258 training frames, 2419 validation frames, and
2419 testing frames. Our final dataset consisted of roughly eight data/image frame pairs per second.

\subsection{Software Tools}
\label{sim}
We needed a way to visually verify that the steering, brake, and throttle values (predicted and actual) reasonably match the cone alignment in an given image frame. This was important to validate that our data/image frame synchronization was done correctly. We juxtaposed a visual of the vehicle data with the frame (Figure ~\ref{fig:intro}), and stitched these composite frames together to make a video rendering all steering, brake, and throttle decisions along the track [4]. This simulator also helped us visualize the output of our model and get a more intuitive grasp of the performance of our models.

All network definitions and training were originally done in Tensorflow, but because of compute and memory limitations on the author's PC, we switched to MIT's OpenMind GPU cluster (details in the acknowledgements). The cluster only supported Torch, so all models, training/testing scripts were translated to Lua/Torch and trained on this cluster. All our Lua and data pre-processing code is available on Github \cite{github1, github2}.

\section{Model-Learning and Experiments}
\label{experiments}
\subsection{Discretized Steering: Model Architecture and Training}
Our first set of models and experiments investigated the feasibility of using CNNs to predict steering direction. Given that our training dataset was less than 0.6\% the size of that used by Bojarski et al. \cite{nvidia}, we were concerned whether it would be possible to learn a model that was able to infer any information about the necessary steering angle at the current position along the track. So, we first decided to tackle a simpler problem to gauge the feasibility of a complete CNN-based steering
controller. We categorized data frames into three groups, frames associated with steering 'left' (\texttt{10 < steering}), steering 'straight' (\texttt{-10 <= steering <= 10}), and steering 'right' (\texttt{steering < -10}). This reduced our controller's job to a classification task: given a snapshot along the racetrack, decide whether to steer left, straight, or right.

After discretizing and one-hot encoding the steering values, we experimented with various CNN architectures to maximize classification performance on the transformed dataset. Our various model architectures are as follows (parenthesized hyper-parameters follow the order listed in the first list item): 

\begin{itemize}
    \item \textbf{1CL-1FC}: 1 Convolutional Layer (CL) (filter sizes: \{$5 \times 5$\}, stride lengths: $\{2\}$, output depths: $\{8\}$). 1 Fully Connected (FC) Layer from reshaped CL output to a (three-node) output.
    \item \textbf{2CL-1FC}: 2 CLs ($\{5 \times 5, 5 \times 5\}$, $\{2,2\}$, $\{8, 16\}$). 1 FC Layer, same as \textit{1CL-1FC}.
    \item \textbf{1CL-2FC}: 1 CL with same hyper-parameters as \textit{1CL-1FC}. 2 FC layers, 100-node hidden layer.
    \item \textbf{2CL-2FC}: 2 CLs with same hyper-parameters as \textit{2CL-1FC}. 2 FC layers, 100-node hidden layer.
    \item \textbf{3CL-2FC}: 3 CLs ($\{5 \times 5, 5 \times 5, 3 \times 3\}$, $\{2,2,1\}$, $\{8, 16, 32\}$). 2 FC layers, 100-node hidden layer.
\end{itemize}

Additionally, we have the following remarks regarding our hyper-parameter tuning and model selection:
\begin{itemize}
    \item We used mean batch normalization after every convolutional layer to control internal covariant shift of values across layers (\texttt{nn.SpatialBatchNormalization}) \cite{batchnorm}. This was added after witnessing very poor validation accuracies for large networks (70-80\% for \textit{3CL-2FC} and larger networks) with training epoch counts less than 200. This oft-used deep-CNN technique we believe was omitted from Bojarski et al. \cite{nvidia}, and was
        necessary for us
        to obtain the low error results we present in section \ref{discresults}. 
    \item We used a final \texttt{softmax} layer to discretize the three-node network output into a one-hot encoding. As is common with classification networks, we used cross-entropy loss and ReLU activations between fully-connected layers (when \texttt{|FC|} > 2).
    \item We borrowed many reasonable hyper-parameter settings (filter-size, stride length, and pooling size) from \cite{nvidia} and models we trained in previous 6.867 assignments \cite{hw3}. Small experiments involving hyper-parameter modification (increasing filter size from 5 to 7, changing stride length from 3 to 1, etc.) yielded few noticeable effects on validation accuracies. Pooling layers were not in architecture prescribed by \cite{nvidia}, but required for us to avoid accuracy
        discrepencies between our training and validation datasets (<5\%). Our pyramidal architecture matches conventional CNN philosophy and is explained in \cite{hw3}. We provide a more rigorous treatment of hyper-parameter optimization for our real-value steering models in section \ref{realsteeringresults}.
    \item We initialized training of all models with a learning rate of 0.01, with an exponential decay of $\frac{1}{1.01}$ every epoch. Learning rate was not discussed in \cite{nvidia}, so we do not have a reference for comparison, but this choice seemed to work relatively well.
    \item All models were trained with batch SGD with batch size 64. Due to memory constraints and I/O bottlenecks on instances on the cluster, we also evaluated validation and testing accuracy in batches of size 64 as well. We report the mean accuracy across the testing and validation batches, evaluated after the model has finished training. All models were trained for 100 epochs (every epoch cycling through all training batches). Unfortunately, \cite{nvidia} does not mention their chosen loss optimizer.
\end{itemize}
As suggested in \cite{nvidia}, the intuition behind multiple CL layers followed by FC layers is to first capture relevant high level features from the incoming video frames, and then feed that information into a "controller" that is the fully connected layers.

\subsection{Discretized Steering: Experimental Results}
\label{discresults}

\begin{figure}
  \begin{minipage}{\textwidth}
    \begin{minipage}{0.5\textwidth}
        \centering
        \includegraphics[width=3in]{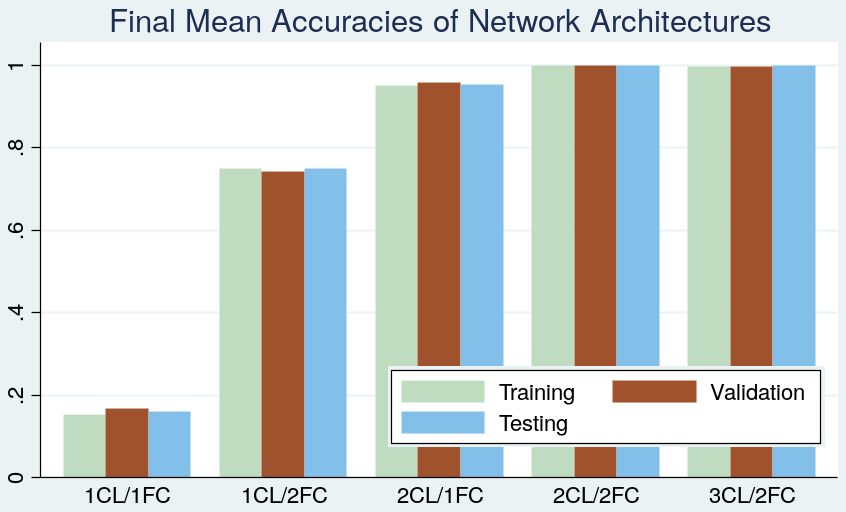}
        \caption{\emph{Accuracies for model architectures. From left to right: 1CL-1FC, 1CL-2FC, 2CL-1FC, 2CL-2FC, and 3CL-FC.}}
        \label{fig:disctot}
    \end{minipage}
  \hfill
  \begin{minipage}{0.5\textwidth}
      \centering
      \includegraphics[width=2in]{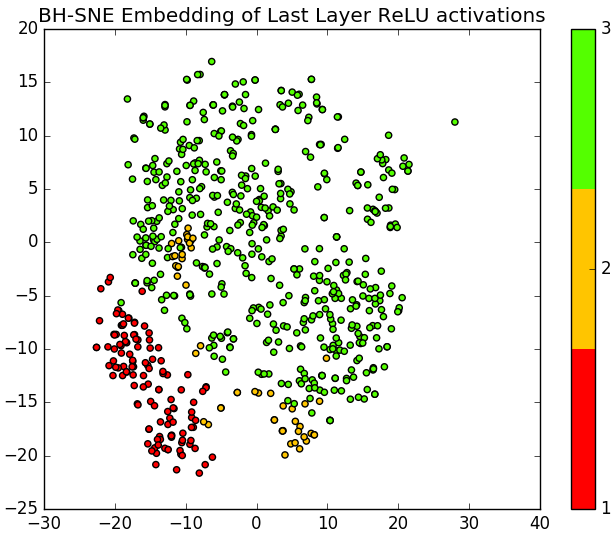}
        \caption{\emph{BH-SNE embedding of the last layer ReLU activations. 1 (red) corresponds to steer 'left' frames, 2 (yellow) is 'straight', and 3 (green) is 'right'.}}
      \label{fig:tsne}
  \end{minipage}
  \end{minipage}
\end{figure}

% \begin{wrapfigure}{H}{0.3\textwidth}
%   \centering
%   \includegraphics[width=2in]{disc_tot.png}
%   \caption{\emph{Accuracies for model architectures. From left to right: 1CL-1FC, 1CL-2FC, 2CL-1FC, 2CL-2FC, and 3CL-FC.}}
%   \label{fig:disctot}
% \end{wrapfigure}

We found only \textbf{2CL-2FC} and \textbf{3CL-2FC} to be successful on our testing datasets after training. The final accuracy results are shown in Figure~\ref{fig:disctot}. Not surprisingly, deeper models tend to perform better: the 2CL-2FL and 3CL-2FL models perform nearly equivalently with both exhibiting 99.7\% and 99.8\% validation and test accuracies, respectively. One convolutional layer was insufficient to extract meaning from data frames; the two models with 1CL exhibited less than
80\% testing accuracy.

Interestingly, and perhaps by random initialization fluke, the 2CL-2FC network exhibited numeric instability problems while training and it took two 100-epoch attempts to finally achieve the accuracies shown in Fig~\ref{fig:disctot}. We noticed a precipitous drop of 60\% in accuracy around training epoch 30, suggesting some sort of zero-propagation/numeric instability problem. For space reasons, we omit the graph here and refer the reader to \cite{instability}.

% \begin{wrapfigure}{H}{3in}
%   \centering
%   \includegraphics[width=3in]{tsne.png}
%     \caption{\emph{BH-SNE embedding of the last layer ReLU activations. 1 (red) corresponds to steer 'left' frames, 2 (yellow) is 'straight', and 3 (green) is 'right'.}}
%   \label{fig:tsne}
% \end{wrapfigure}

In order to better understand the learned network's behavior, we extracted final round activations for two batches of test data and used a Barnes-Hut SNE embedding from 100-dimensional space into two dimensions, which we show in Fig~\ref{fig:tsne}. Last-layer activations that have the same label coalesce with a pleasing topology: steer-'left' frames cluster on the left, steer-'straight' frames cluster in the center, and steer-'right' frames cluster on the right. As is suggested by the
plot, the dataset is lopsided, with many more right turns than left turns or straightaways (a characteristic of the racetrack). Furthermore, it seems as though the network has trouble distinguishing straight frames from right and left frames. This is confirmed by the test dataset confusion matrix: of the five misclassified testing frames, all five had true or predicted label 'straight'. This is likely because the steering cutoff values we chose for three-bin discretization (-10 and 10), albeit reasonable, were arbitrary, and true steering values are on a continuous range from -90 to 90.

\subsection{Real-value Steering: Model Architecture and Training}
\label{steerresults}
A complete controller is not only able to provide high-level, discretized steering commands, but also output the angle of the steering wheel. We extend our controller to output a real-value steering angle given a video frame along the track. Our networks closely match those we made for discretized steering. \cite{nvidia} tackles this regression task with a much deeper network (five convolutional and three fully connected layers), so starting with our best performing network from \ref{discresults} (3CL-2FC), we make a few modifications to optimize for real value steering:
\begin{itemize}
    \item Our output layer is no longer a \texttt{softmax}, rather a one-node output that feeds into Torch's \texttt{nn.HardTanh}, an \texttt{tanh} non-linearity which we scale to $[-90, 90]$ to match the range of possible steering values.
    \item Cross-entropy loss is not appropropiate for our output range, so we applied smooth L1 error as our loss criterion (\texttt{nn.smoothL1Criterion}). Like before, we used Torch's batched SGD optimizer. We chose L1 loss, as compared to MSE or similar real-valued criterion, because it is easily interpretable and more robust to outliers \cite{l1}, which is likely to occur in our data because of FTDI/UART signal corruption and fickle steering-axis potentiometers.
\end{itemize}
We trained three different deeper models architectures: \textbf{3CL-2FC}, \textbf{3CL-3FC}, and \textbf{4CL-3FC}. The two hidden layers in 3CL-3FC and 4CL-3FC had layer sizes of 1024 and 100, in that order. The 4CL-3FC contained the same convolutional hyper-parameters as the 3CL networks for the first three CLs, with an extra $3 \times 3$, stride 1 filter, depth 48 for the last CL. We demonstrate small improvements tuning hyper-parameters via grid-search for the strongest model \textbf{4CL-3FC}
as shown in the next section.

\subsection{Real-value Steering: Experimental Results}
\label{realsteeringresults}
\begin{wrapfigure}{H}{2in}
  \centering
  \includegraphics[width=2in]{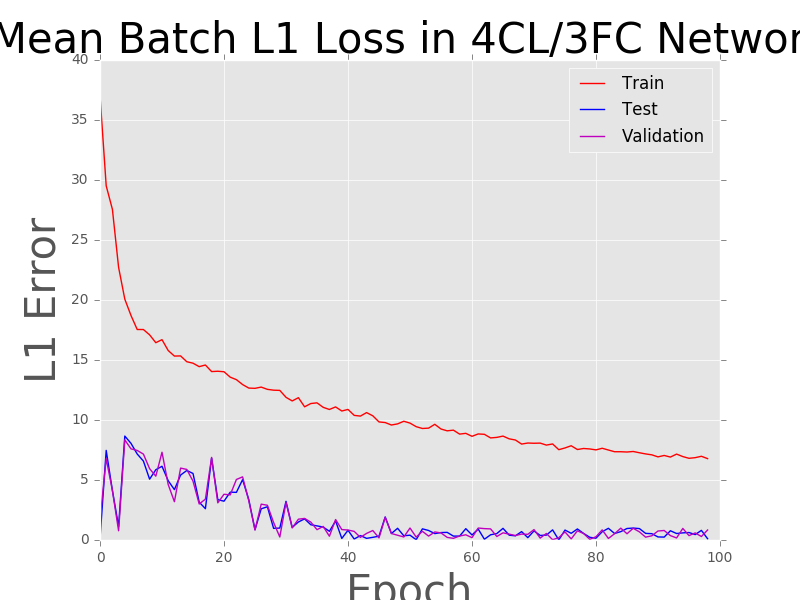}
    \caption{\emph{Change in L1 loss across training epochs for 4CL-3FC. Training (red) upper bounds validation (magenta) and testing (blue).}}
  \label{fig:real}
\end{wrapfigure}

The mean absolute deviations across all frames in our training, testing, and validation datasets are given in Table~\ref{tab:realtot}. Interestingly, \textbf{4CL-3FC} is the strongest network on the testing dataset (with a mean deviation of 0.1 steering degrees from a model driver), but is \textit{not} the strongest network on either the training or validation datasets. Furthermore, the loss on the training dataset is significantly larger than either the validation or testing loss (see Figure
~\ref{fig:real}). We draw two conclusions from this:
\begin{itemize}
    \item The significantly larger error on the training dataset, not reflected in either the validation or testing datasets, likely suggests that there are outliers in the training dataset (either erroneous steering values or situations on the track in which the steering angle may deviate significantly.) Because the training dataset is larger than the validation and testing datasets, it could be more susceptible to low-occurence erroneous values. That we did not witness such
        results in our discretized testing suggests that the outlier(s) were consistent in steering direction, but anomalous with respect to magnitude.
    \item Interestingly, the deeper network 4CL-3FC performs worse on the training, but comparatively better on validation and significantly better on testing. We were able to rule out random initialization as the cause by witnessing the behavior even after re-training. This suggests there is some sort of regularization effect in the deeper network. We conjecture that this is caused by the extra layer of max-pooling which halves the layer area -- forcing a sparser, more high-level representation
        of the current driving state. This helps regularize 4CL-3FC.
\end{itemize}

\begin{wraptable}{b}{3in}
    \label{tab:realtot}
    \begin{tabular}{lllll}
        \textbf{}                                & \textbf{4CL-3FC}             & \textbf{3CL-3FC}             & \textbf{3CL-2FC}             &  \\ \cline{2-4}
        \multicolumn{1}{l|}{\textbf{Training}}   & \multicolumn{1}{l|}{6.78182} & \multicolumn{1}{l|}{5.74226} & \multicolumn{1}{l|}{6.3078}  &  \\ \cline{2-4}
        \multicolumn{1}{l|}{\textbf{Validation}} & \multicolumn{1}{l|}{0.83272} & \multicolumn{1}{l|}{0.96716} & \multicolumn{1}{l|}{0.5236}  &  \\ \cline{2-4}
        \multicolumn{1}{l|}{\textbf{Test}}       & \multicolumn{1}{l|}{0.10480} & \multicolumn{1}{l|}{0.89492} & \multicolumn{1}{l|}{0.87435} &  \\ \cline{2-4}
    \end{tabular}
    \caption{Mean L1 Errors of Model Architectures}\label{wrap-tab:1}
\end{wraptable}

\begin{wrapfigure}{H}{1.5in}
  \centering
  \includegraphics[width=1.5in]{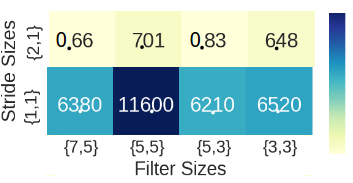}
    \caption{\emph{Grid search across hyper-parameter configurations. Plotted in varying colors is the validation L1 error of the 4CL-3FC network. To reduce our search space, the configurations that we tried were double compressed; for example, filter size '\{5,3\}' corresponds to two $5 \times 5$ filters, and two $3 \times 3$ filters.}}
  \label{fig:hyp}
\end{wrapfigure}

The results of our hyperparameter grid search across various filter sizes and stride sizes are shown in Fig~\ref{fig:hyp}. The improvement in validation error from tuning hyper-parameters from our original values to their optimal values is small, yielding a 0.2 drop in L1 validation error, achieved with filter sizes of $\{7,7,5,5\}$ and stride lengths of of $\{2,2,1,1\}$. For time reasons, we did conduct experiments across max-pooling sizes, but our experiments in \cite{hw3} informed our selected
pooling sizes. Matching practices in \cite{nvidia}, we chose not include dropout in our network. The negative validation and training error difference suggests we are not overfitting, and do not need further regularization.

Using the simulator described in \ref{sim}, we visualized our network's predicted steering angle for a continuous series of frames during a track run. The results can be seen on YouTube \cite{steer_vid}. From this, we learned a few things. First, the controller learns a highly non-continuous behavior because each frame is independent from the previous frame. The model borders on erratic, and is not yet suitable to be placed in a car, even though it apparently has a low deviation from a
model driver. While it seems as though the general steering direction is mostly correct, the model's scaled final activation (\texttt{HardTanh}) seems to reach its borders frequently (jumping the steering to hard left or hard right). Perhaps for future models, we may penalize large discontinuities in predicted steering, and introduce recurrency to the model so that frames are not independent. 

On a frame by frame basis, the network by-and-large seems to produce plausible steering angles, but when stitched together, the behavior is erratic, similar to other driverless vehicle demonstrations \cite{stanford}. It is also possible that we did not correctly synchronize the frames and the networks predicted steering, which could also explain the network's erratic frame by frame behavior.

We rendered some of the learned convolutional filters to understand network behavior; Fig~\ref{fig:llfilts} illustrates four of our last layer filters. It was hard for us to draw meaning from the filters geometries, but they do seem to be very structured, and at the very least, picking up on general track trajectories. For space reasons, we omit printing all filters here, but the results are publically viewable at \cite{convfilt}.

On a 64GB RAM/Intel Core i5 (2.7GhZ) computing instance with a NVIDIA Tesla M60 GPU, the average pass-through latency of frame to steering angle was 0.22 milliseconds. On average, every convolutional layer in our network added 0.02 ms to the latency, while every additional fully connected layer added 0.01 ms to the latency. For space reasons, we cannot provide the breakdown, but we provide our logs and analysis scripts to the interested reader \cite{slurm}.

\subsection{Shift-Translation Data Augmentation}
\label{aug}
\begin{wrapfigure}{H}{2in}
  \centering
  \includegraphics[width=2in]{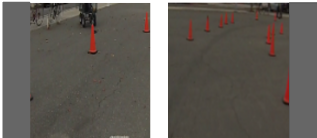}
    \caption{\emph{Left: example of a frame shift right. Right: example of a frame shift left.}}
  \label{fig:shifts}
\end{wrapfigure}

One common failure mode of controllers trained on data from expert drivers is non-model circumstances. If the network makes a small small error and veers slightly left or right, it may result in a situation not thoroughly trained by model driving data. The network may potentially make even poorer decisions with little certainty about its situation, further compounding the errors until the vehicle crashes.

An approach we tried to tackle this issue that we tried implementing was data augmentation by way of image translation (Fig~\ref{fig:shifts}). Should the vehicle deviate from center, the forward facing camera(s) experience a perspective shift that can be roughly modeled by an image translation. We labeled this shifted frame dataset with a corresponding shift in steering angle proportional to the degree to which we shifted the image (filling the empty space in the frame with the mean pixel value).
Training on a non-augmented dataset and testing on a mixed 15\% normal/85\% shifted set resulted in a mean L1 validation loss of 18.1. Including translated frames in our training decreased the validation error to 7.4 but increased the non-augmented testing data set error to 1.4. Without an accurate driving simulator, it is hard to conclusively judge the effectiveness of data augmentation, but at the very least, the idea of translation-based augmentation is supported by \cite{nvidia}.

\subsection{Preliminary Work on Brake/Throttle Controllers}
Extending our work with real-value, we aimed to develop similar controllers for brake and throttle. Developing and testing our steering models required over 35 hours, and unfortunately, due to time constraints, the author is not able to demonstrate validation and testing results for this network. We do however, describe our Torch implementation and the architecture of our model as we currently are building it.

\begin{figure}
  \begin{minipage}{\textwidth}
    \begin{minipage}{0.5\textwidth}
      \centering
      \includegraphics[width=4in]{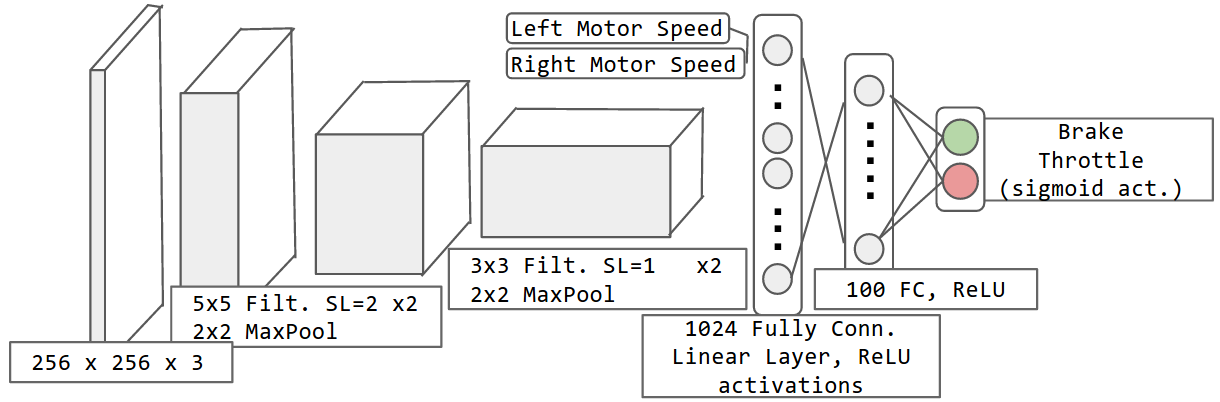}
        \caption{\emph{Proposed architecture for the brake/throttle controller network.}}
      \label{fig:brake}
    \end{minipage}
  \hfill
  \begin{minipage}{0.5\textwidth}
      \centering
      \includegraphics[width=2in]{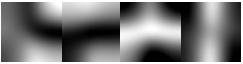}
        \caption{\emph{Four last layer convolutional filters.}}
      \label{fig:llfilts}
  \end{minipage}
  \end{minipage}
\end{figure}

% \begin{wrapfigure}{H}{4in}
%   \centering
%   \includegraphics[width=4in]{brake_net.png}
%     \caption{\emph{Proposed architecture for the brake/throttle controller network.}}
%   \label{fig:brake}
% \end{wrapfigure}

The network layout, based on our experiments for steering, are illustrated in Fig~\ref{fig:brake}. Importantly, video feed is insufficient for making a decision on brake and throttle -- the current state of motion, captured by the current motor speeds, are necessary as well. We directly feed these into the fully connected 'controller' layers. Furthermore, because brake and throttle range from 0 to 256, we use a scaled sigmoid on our two-node output. The network is not a straight line
path, rather a multi-parent DAG, so we had to import extra Torch library, \texttt{nn.graph}, in our implementation \cite{nngraph}. Note that the architecture and hyper-parameter values are likely to change based on future experimentation and comparison of validation errors.

% \begin{wrapfigure}{H}{2in}
%   \centering
%   \includegraphics[width=2in]{ll_filters.png}
%     \caption{\emph{Four example last layer convolutional filters.}}
%   \label{fig:llfilts}
% \end{wrapfigure}

\section{Conclusion}
\label{conclusion}
In this report, we have presented a set of CNN-based end-to-end models for steering, along with various benchmarking and visualization tools to understand their performance. We tackled three main problems in the context of cone-delineated racetrack driving: (1) discretized steering, which translates a first-person frame along to the track to a predicted steering direction. (2) real-value steering, which translates a frame view to a real-value steering angle, and (3) a network design for
predicting brake and throttle. We demonstrate high accuracy on our discretization task, low theoretical testing errors with our model for real-value steering, and a starting point for future work regarding a controller for our vehicle's brake and throttle. Timing benchmarks suggests that the networks we propose have the latency and throughput required for real-time controllers, when run on GPU-enabled hardware.

Translating this to the MIT Motorsports racecar will require more work to correct failure modes that we've identified: as shown, basic simulation suggests that frame-by-frame steering, as employed by \cite{nvidia}, is insufficient to produce smooth steering behaviors (or incompatible with our models). Also, more extensive data augmentation and responsive simulation would be required to avoid the "compounding error" problem mentioned in section \ref{aug}. Recurrency, probabilistic
methods, classical path-planning and localization algorithms, and end-to-end safety checks are all necessary before we roll out a driverless racecar for FSGD.

\subsubsection*{Acknowledgments}
I’d like to thank Kevin Chan, Michael Janner, and Luis Mora, three teammates on the MIT Motorsports team, for invaluable help making modifications to the racecar and collecting data. Special thanks to Michael for forwarding Torch jobs to the OpenMind MIT Brain and Cognitive Sciences Cluster and returning the job's logs and output to me. Acknowledgement to Olivia Zhao for working through graphing challenges in \texttt{Stata} with me. I’d also like to thank Jonathan Huggins and Professor Lozano-Perez for directing me to relevant papers and providing me with guidance on model setup and training.

\medskip

\tiny

\end{document}